\documentclass{article}
\usepackage{amsmath,graphicx,mlspconf}
\usepackage{bm}
\usepackage{amssymb}
\usepackage{amsfonts}

\usepackage{color}
\usepackage{times}
\usepackage{epsfig}
\usepackage{epstopdf}
\usepackage{caption}
\usepackage{graphicx}
\usepackage{amsmath}
\usepackage{amssymb}
\usepackage{subfigure}
\usepackage{latexsym}
\usepackage{dblfloatfix}

\usepackage[linesnumbered, ruled, vlined]{algorithm2e}  
\usepackage[capitalise]{cleveref}                       

\usepackage{todo}

\graphicspath{{images/}}

\title{Limiting the Reconstruction Capability of Generative Neural Network using Negative Learning}
\name{Asim Munawar, Phongtharin Vinayavekhin and Giovanni De Magistris}
\address{IBM Research - Tokyo}
\begin{document}
%

\maketitle
%
\begin{abstract}
  Generative models are widely used for unsupervised learning with various
  applications, including data compression and signal restoration.
  Training methods for such systems focus on the generality of the network given limited amount of training data.
  A less researched type of techniques concerns generation of only a single type of input.
  This is useful for applications such as constraint handling, noise reduction and anomaly detection.
  In this paper we present a technique to limit the generative capability of the network using negative learning.
  The proposed method searches the solution in the gradient direction for the desired input and in the opposite direction for the undesired input.
  One of the application can be anomaly detection where the undesired inputs are the anomalous data.
  In the results section we demonstrate the features of the algorithm using
  MNIST handwritten digit dataset and latter apply the technique to a real-world obstacle detection problem.
  The results clearly show that the proposed learning technique can significantly improve the performance for anomaly detection.  

\end{abstract}

\section{Introduction}
\label{sec:intro}

Generative networks can learn to generate high-dimensional data from lower dimensional embeddings.
Most of the applications require the generative models to generalize given a limited amount of training data.
As a consequence, even the signals that are far from the training data distribution can be generated fairly well.
Controlling this generalization property of the generative networks can increase their efficiency in the domains where we need to separate one kind of data from the other.
Some of the applications of system with limited generative ability include noise reduction and anomaly detection.

In this paper, we present a method to control the generative capabilities of a system in such a way that it can only reconstruct a limited range of input signal space.
The technique can be used with different network structures and training algorithms.
We will explain the proposed method by focusing on anomaly detection in higher dimensional spaces (e.g. images etc.) using a kind of generative neural networks called autoencoder.
Using the proposed technique, generative models can be trained in a way to learn a latent representation that can only encode the input distributions of non-anomalous data.
After decoding the latent space back to the signal space, the reconstruction similarity can be used to judge if the input signal contains an anomaly or not.

Anomaly detection is a key and usually the first requirement in many signal-processing applications pipeline~\cite{forslundIV2014, munawar2017}.
Generative models have previously been applied to anomaly detection~\cite{sakurada_2014, clement2015} and noise reduction~\cite{VincentPLarochelleH2008}.
In anomaly detection the task is to find if the input distribution is normal or has an anomaly.
It is a one-class classification problem where the training data consists mostly of the non-anomalous class.
We argue that due to the generalization property the classic training methods are not ideal for anomaly detection using generative models (as shown in Section \ref{sec:results}).

The main contribution of this paper is a new approach to limit the reconstruction capability of the generative networks by learning conflicting objectives for the normal and anomaly data.
The technique can use the limited real or synthetic anomalous data by using a negative learning phase in the training.
For example, in case of anomaly detection on the road~\cite{clement2015}, any non-road object (e.g. vehicles, bushes etc.) can be treated as the anomalous data.
Some anomaly data is available in most of the anomaly detection applications.
The anomaly data might be gathered over time automatically or by human intervention.
For instance, in case of a misclassification by a radar based target detection system, the human operator can label the sample correctly for future use.
Instead of ignoring this anomaly data, the proposed method uses this data to improve the future detections. 

The rest of the paper is organized as follows. Related work is given in Section~\ref{sec:relatedwork}.
We formally define the problem in Section~\ref{sec:ps}.
The specificities of our approach are detailed in Section~\ref{sec:details}.
Quantitative analysis of the technique are presented in Section~\ref{sec:results}.
Finally, we conclude the paper with some directions for future work in Section~\ref{sec:conclusions}.

\section{Related Work}
\label{sec:relatedwork}

There are a large number of literatures on noise reduction and anomaly detection using generative models.
M.N. Schmidt et al. \cite{schmidt2007} uses non-negative sparse coding to reduce the wind noise in speech data.
They rely on a system that have the source model for the wind noise but not for the speech to reduce the noise.
The work done on denoising autoencoder by V. Pascal et al.~\cite{pascal2010} is also very important in this area.
L. Gondara~\cite{gondara2016} presents an application of such denoising system to remove noise from medical images.
These techniques can reduce the noise the input data but they do not limit the generative capabilities of the network.
Due to the generalization property of such networks, they can also generate the data that is very different from the data shown during training.



Similar to the method proposed in this paper, for anomaly detection, the machine learns a model to represent normality and then use the model to detect anomalous data.
B. Saleh et al.~\cite{SalehCVPR2013} proposed a method to model a normality of a particular class of object using visual attributes.
The attributes~\cite{FarhadiCVPR2009} are handcrafted and mainly based on the appearance of the input data, i.e. shape, texture and color.
A generative model is then trained and used to reason about normal and anomalous data.
Recent trend tends to replace these handcrafted attributes with a deep feature representation.
W. Lawson et al.~\cite{Lawson_2016_CVPR_Workshops} uses deep visual features obtained from AlexNet~\cite{alexnet} to represent objects and associated them with a scene to define type of objects that can be found in the certain environment.
D. Xu et al.~\cite{xu2015bmvc} used stacked denoising autoencoders to learn the deep features in an unsupervised fashion and use them to represent both appearance and motion of the scene.
Anomalous data is in turn detected by a multiple one-class SVM classifiers.
These approaches are more likely to suffer from the imbalance between normal and anomalous data which are the common characteristic of an anomaly detection problem,
The proposed method try to solve this problem by effectively using the anomaly data.

Our proposed method uses a similar approach to C. Creusot and A. Munawar~\cite{clement2015}.
They use an extremely compressive Restricted Boltzmann Machine (RBM) to form a deep feature representation.
But rather than training a classifier in the feature space, anomaly detection is performed by reconstructing the data back to the original image space and use conventional image difference as a metric.
The extreme compression in autoencoders can severely effect the reconstruction of input appearance in case the non-anomalous data have a non-trivial appearance.

\section{Problem Statement}
\label{sec:ps}

In this section, we will formally describe the problem of limiting the generative network to learn a single type of input distribution.
Consider two random variables $\textbf{X}$ and $\textbf{Y}$ representing instances of two input distributions in same signal space (e.g. image space).
Lets assume we have $K$ and $J$ number of samples from each distribution, $\textbf{X}=\{\textbf{x}_1,...,\textbf{x}_K\}$ and $\textbf{Y}=\{\textbf{y}_1,...,\textbf{y}_J\}$.
$\textbf{X}$ is the input distribution we want the network to reconstruct as well as possible, let the reconstruction be called $\hat{\textbf{X}}$.
On the other hand, $\textbf{Y}$ is the distribution that we do not want to the network to reconstruct. Let its reconstructed space be represented by $\hat{\textbf{Y}}$.
In order to achieve this objective, we need to maximize

\begin{equation}
  \label{eq:x}
  p_{\theta }(\hat{\textbf{X}}|\textbf{X})-p_{\theta }(\hat{\textbf{Y}}|\textbf{Y}) = \sum_{i=1}^K \log{p_{\theta}(\hat{\textbf{x}}_i|\textbf{x}_i)}-\sum_{i=1}^J \log{p_{\theta}(\hat{\textbf{y}}_i|\textbf{y}_i)}
\end{equation}

\noindent By maximizing the probability of reconstruction for $\textbf{X}$ and minimizing it for $\textbf{Y}$, the generative properties of the model can be controlled in the desired way.
It is important to note that usually the data for the distribution $\textbf{X}$ is available in plenty while the data for $\textbf{Y}$ is available scarcely ($K\gg J$).

\section{Proposed Method}
\label{sec:details}

In this section we discuss the proposed approach to maximize Equation~\ref{eq:x}.
Generative models can be used in a variety of settings and configurations.
In this paper we deal with the generative models that encode the input distribution into a latent feature space $L$ and then reconstruct it back in the original signal space.
Such generative systems are also known as autoencoders.
We use the word ``autoencoder'' for any kind of generative neural network structure including but not limited to RBM, variational autoencoders and Convolutional Neural Network based autoencoders.

The problem is to learn latent representation $\textbf{L}$ such that it can learn to encode and decode $\textbf{X}$ fairly well but fails to do the same for $\textbf{Y}$ distribution.
In order to formally define the autoencoder like generative models, let us consider a network with input vector of size $N$, a latent space or hidden layer of size $H$.
As the network will learn to reconstruct the input the output of the network will also be of size $N$.
Given the training data $\textbf{X}$, a function $\textbf{F}$ can transform this input signal to the hidden layer while a function $\textbf{G}$ can reconstruct the image from the latent space.
The network parameters for encoder and decoder are represented by $\alpha$ and $\beta$ respectively.
We want to find the optimal parameters $\theta=\{\alpha, \beta\}$ to minimize the reconstruction error.
When presented with an input vector $\textbf{x}$, the network produces a hidden vector $\ell=\textbf{F}(\textbf{x};\alpha)$ and an output vector $\hat{\textbf{x}}=\textbf{G}(\ell;\beta)$.
The goal of the learning is to minimize an error or energy function $E$

\begin{equation}
  \begin{aligned}
    \textup{min }E(\textbf{F},\textbf{G}) &= \underset{\textbf{F},\textbf{G}}{\textup{min}}\sum_{i=1}^{K}\Delta(\hat{\textbf{x}}_i,\textbf{x}_i) \\
                                          &= \underset{\textbf{F},\textbf{G}}{\textup{min}}\sum_{i=1}^{K}\Delta(\textbf{G}(\textbf{F}(\textbf{x}_i;\alpha);\beta)), \textbf{x}_i)
  \end{aligned}
\end{equation}

\noindent where, $\Delta$ is a distance or dissimilarity measure. We can use any dissimilarity measure, in this paper we use mean square error: $\Delta=\sum_{i=1}^K{(\hat{\textbf{x}_i} - \textbf{x}_i)}^2$. Optimum set of parameters $\theta^*$ can be found by

\begin{equation}
  \label{eq:pos}
  \theta^*=\underset{\theta}{\textup{arg min}} \sum_{i=1}^K{(\hat{\textbf{x}_i} - \textbf{x}_i)}^2
\end{equation}
In order to create an interesting representation of the data, usually the size of hidden layer is kept smaller then the input size $H<N$.
However, $H\geq N$ can also be used with additional sparsity constraints to see very interesting behaviors for some applications.

\begin{algorithm}[t]
	\caption{Training algorithm with a negative learning phase.}
    \label{alg:flow}
    \KwData{$\textbf{X}$; Non-anomalous data.}
    \KwData{$\textbf{Y}$; Anomalous data.}
    \KwIn{$Q$; Number of iteration of a negative learning phase.}
    \KwResult{An autoencoder that can reconstruct $\textbf{X}$ properly, but poorly on $\textbf{Y}$.}
	\DontPrintSemicolon\
	|||START|||\;
	\tcc{Perform a positive learning with $\textbf{X}$.}
	\sc{PositiveLearning}($\textbf{X}$)\;
	\tcc{Loop until the termination criteria is satisfied.}
	\While{Not Satisfying the Termination Criteria }{
                \tcc{Multiple iterations of negative learning are performed; in order to balance the ratio of $\textbf{X}$ and $\textbf{Y}$.}
		\For{$i=0$ \KwTo $Q$}{
			\tcc{Perform a negative learning with $\textbf{Y}$.}
			\sc{NegativeLearning}($\textbf{Y}$)\;
		}
		\tcc{Perform a positive learning with $\textbf{X}$.}
		\sc{PositiveLearning}($\textbf{X}$)\;
	}
	|||END|||\;
\end{algorithm}

In this paper, we propose using any real or synthetic anomalous data $\textbf{Y}$ to limit the reconstruction capability of the autoencoder.
This is done by incorporating a negative learning phase in the training.
System parameters learned during the training allow reconstructing a wide variety of input patterns.
Negative training adjusts the system parameters in a way that the anomalous patterns cannot be reconstructed well.
In terms of neural networks, the connections that are used to reconstruct anomalies are weakened during the negative learning.
The negative learning can formally be defined as

\begin{equation}
  \label{eq:neg}
  \theta^*=\underset{\theta}{\textup{arg max}} \sum_{i=1}^J{(\hat{\textbf{y}_i} - \textbf{y}_i)}^2
\end{equation}

Using Equation \ref{eq:pos} and Equation \ref{eq:neg}, the model can be controlled to reconstruct non-anomalous data better than the anomalies.
It is important to note that both the equations are optimizing the same set of parameters $\theta$ with conflicting objectives.
Going along the gradient for non-anomalous data and against the gradient for negative makes the system go towards a minimum where it can reconstruct only non-anomalous data.

Algorithm~\ref{alg:flow} introduces the negative learning phase.
The strategy is to use all the non-anomalous data and finish one epoch of positive learning in which the system learns to reconstruct the non-anomalous training data $\textbf{X}$.
Then in the negative learning step the available anomalous data $\textbf{Y}$ is used to make the system unlearn the reconstruction capability of the anomalies.
Positive and negative learning steps are repeated until the termination criteria is met.
This enables the system to learn only the reconstruction of non-anomalous signals.
We show that the benefits of using the negative learning approach are significant even when the size of anomalous training samples is much smaller than the non-anomalous signal data.

It is important to keep a balance between the negative and the positive learning.
If the size of anomalous data is very small compared to non-anomalous data, a single positive learning iteration should be followed by multiple iterations of negative learning.
An adaptive approach can be used to compute optimal number of iterations for the negative learning.
This adaptive algorithm is out of the scope of this paper.

\setcounter{figure}{1}
\begin{figure*}[b!]
\centering
\subfigure[Frequency distribution curve of the conventional autoencoder.]{\
    \centering
	\includegraphics[width=.3\textwidth]{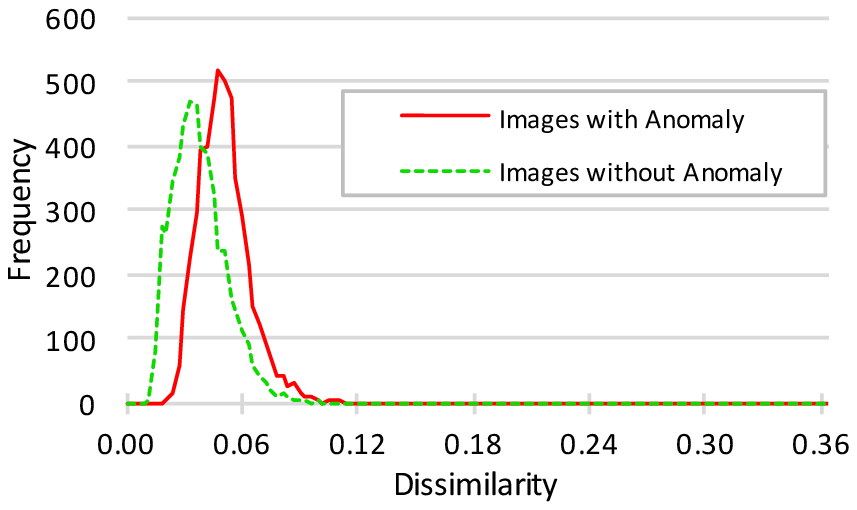}
\label{fig:freq_a}
}
\subfigure[Frequency distribution curve of the proposed autoencoder.]{\
  \centering
  
	\includegraphics[width=.3\textwidth]{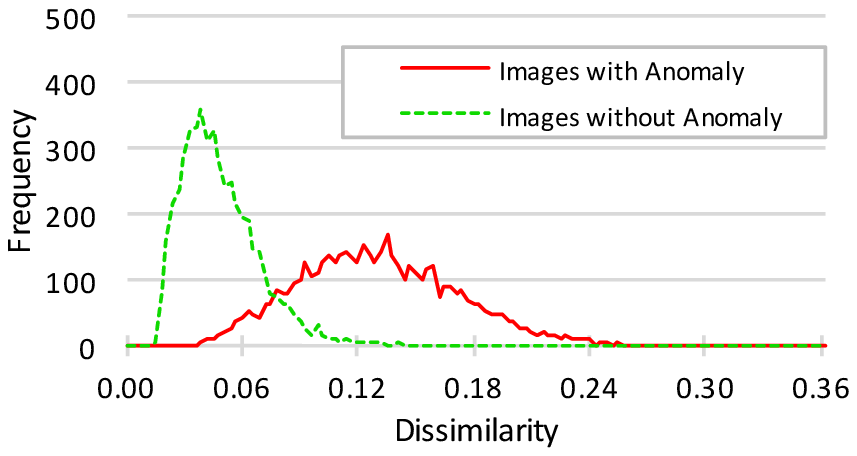}
\label{fig:freq_b}
}
\subfigure[Frequency distribution curve of the proposed autoencoder, when a small amount of anomalous data is used in a negative learning phase.]{\
    \centering
    \includegraphics[width=.3\textwidth]{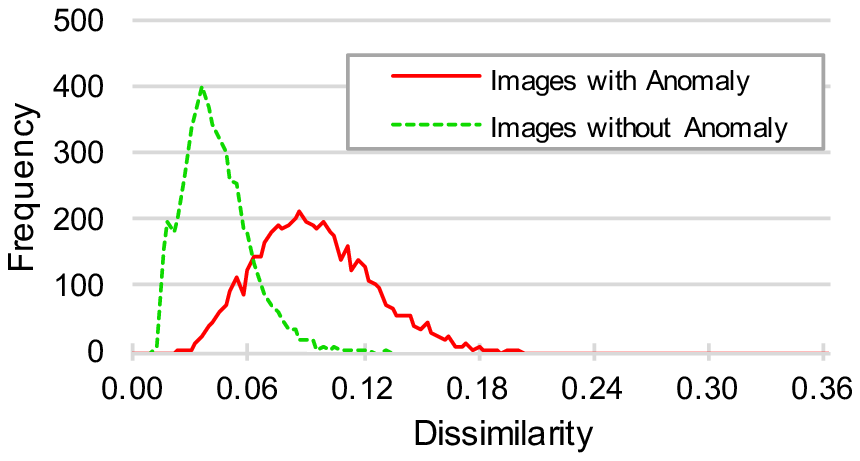}
\label{fig:freq_c}
}
\caption{The dissimilarity between the observed and the reconstructed images; \textit{red} line shows a dissimilarity frequency of anomalous images, and \textit{green} line show a dissimilarity frequency of non-anomalous images. Anomaly detection can perform effectively when these two curves are horizontally far apart (easily vertically separable by a threshold).}
\label{fig:freq}
\end{figure*}

\section{Experimental Results}
\label{sec:results}

In order to explain the working of the algorithm, the initial experiments are conducted with the MNIST digits dataset.
The later part of this section uses actual highway data to show the validity of the approach for real world problems.

\setcounter{figure}{0}
\begin{figure}[t!]
\centering
\includegraphics[width=0.9\columnwidth]{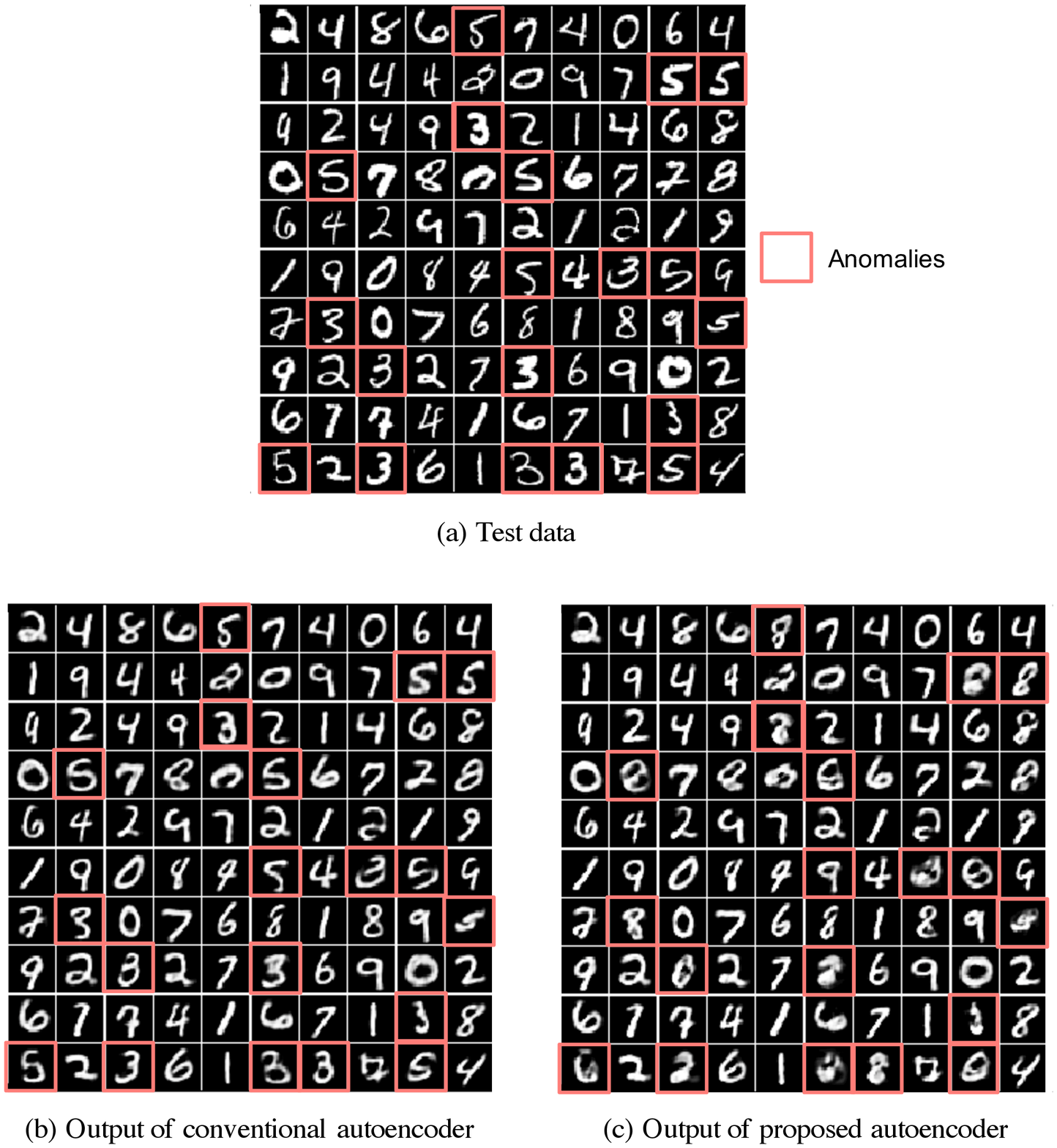}
\caption{Conventional autoencoder can reconstruct any input, but the proposed system fails to reconstruct anomalies (digits $3$ and $5$).}
\label{fig:our_rec}
\end{figure}

\setcounter{figure}{2}

\subsection{Evaluation using MNIST}

For this experiment we have used a single layer RBM based autoencoder.
$28\times28$ gray-scale images of MNIST digit dataset are used to train a fully connected autoencoder of size 784$-$500$-$784.
Sigmoid was used as the activation function.
Termination criteria, maximum number of epochs was set to $200$.
Batch size was $50$.
The network was trained by using single-step contrastive divergence (CD-1)~\cite{Hinton2012}.

\begin{equation}
  \label{eq:cd}
  \Delta w_{ij}=\zeta \left [  \epsilon(\left \langle {v_i}{h_j} \right \rangle_{data}-\left \langle {v_i}{h_j} \right \rangle_{recon}) \right ]
\end{equation}
\noindent where $\mathbf{v}$ represents visible layer, $\mathbf{h}$ is the hidden layer, $\epsilon$ represents the expected value and $\zeta$ is the sign.
For positive learning stage the change in weights are updated by Equation~\ref{eq:cd} with $\zeta=1$, and for negative stage (going against the gradient) the same equation is used with $\zeta=-1$.

Figure~\ref{fig:our_rec}(a) shows images from MNIST test dataset.
Digit $3$ and $5$ are considered to be the anomalies; hence, the autoencoder trained with the proposed method should not be able to reconstruct these digits well.

Figure~\ref{fig:our_rec}(b) shows the results of reconstruction using a conventional autoencoder, trained using CD-1 with $200$ epochs and batch size of $50$.
The training data for conventional training method contains all the digits except the images of digit $3$ and $5$.
It can be clearly seen that that even though the system knows nothing about digits $3$ and $5$, it is able to reconstruct them fairly accurately.
This property of autoencoders is not desirable for anomaly detection.

Figure~\ref{fig:our_rec}(c) shows the reconstruction results using the proposed approach.
Digits $3$ and $5$ are no more reconstructed properly; rather they are converted to the closest point in the non-anomalous signal space.
From the shapes point of view, digits $3$ can be thought of as a part of digit $8$ and similarly digit $5$ is a part of digit $6$.
Yet the system trained with the proposed approach was able to reconstruct digits $8$ and $6$ but failed to reconstruct $3$ and $5$.

Figure~\ref{fig:freq} shows the frequency distribution of dissimilarity measure for normal and anomaly data.
For the conventional autoencoder, we can observe a huge overlap between the curves, making it difficult to select a suitable threshold to decide anomalies.
The proposed autoencoder shifts and spreads the curve of the anomaly data horizontally while the curve for the non-anomalous data largely remain unaffected. 

To simulate the case where the anomaly data is much smaller than the non-anomalous data, another experiment is conducted using only $1,000$ anomalous images (the first $500$ images of $3$ and $5$), and roughly $50,000$ non-anomalous images for training.
In order to create a balance between the positive and negative learning phase, five iterations of negative learning are performed after each positive learning phase (number of iterations for negative learning can be computed adaptively, this however is out of the scope of current paper).
As shown in Figure~\ref{fig:freq}(c), even for this experiment where the anomalous data size is $50$ times smaller then regular data, there is still a major improvement as compared to the results of conventional autoencoder given in Figure~\ref{fig:freq}(a).
Similar results were achieved using other digits as anomaly.

\begin{figure}[!t]
\centering
\includegraphics[width=0.9\columnwidth]{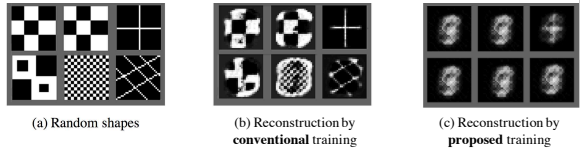}
\caption{Conventionally trained network can encode and reconstruct signals that are very different from the training data (MNIST). The proposed autoencoder failed to reconstruct random shapes even though they were never shown as anomaly data during training.}
\label{fig:randrec}
\end{figure}

Another very interesting results are shown in Figure~\ref{fig:randrec}.
Figure~\ref{fig:randrec}(b) shows that a conventional autoencoder trained solely using the images of digits in MNIST dataset can also reconstruct random shapes.
However, as visible in Figure~\ref{fig:randrec}(c), even though only digit $3$ and $5$ were used as anomalous images during the training process, the system failed to reconstruct anomalies that it has never seen before.
This proves that knowing the appearance of all the anomalies is not important.

\subsection{Evaluation on obstacle detection}

In the second experiment, we used the 4K highway video on Japan highways \cite{youtube1}.
It consists in a 1h40m sequence of Japan highway recorded from the car dashboard with a Panasonic GH4 camera in 4K resolution ($3840\times2160$).
We considered the video between frame~105360 to frame~114360.
\begin{figure}[!h]
\centering
\includegraphics[width=1.0\columnwidth]{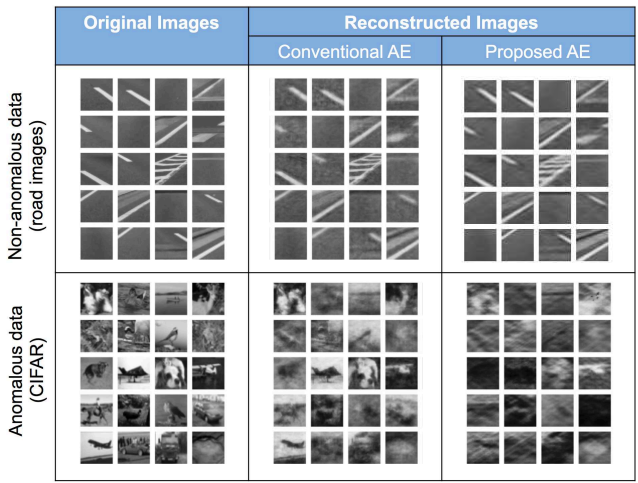}
\caption{Non-anomalous images are reconstructed well with both autoencoders. On the other hands, anomalous images are only reconstructed better on the conventionally trained autoencoders.}
\label{fig:reconstruction_roadcifar}
\end{figure}
These frames were selected as they have a good view of the road without any vehicle occluding the road.
The images were converted to gray-scale and then resized to $25\%$ of the original size.
We then selected a fix mask of $170\times170$ in the center road area.
$24,800$ gray-scale road patches images of size $32\times32$ were extracted with random strides from the rescaled video as non-anomalous data.
During the dataset creating all featureless road patches were ignored.
In this experiment we used just $500$ gray-scale CIFAR-10 images~\cite{krizhevsky2009learning} as anomaly data (as shown in Figure~\ref{fig:reconstruction_roadcifar}).
Randomly selected $70\%$ of the data was used for training while the remaining data was used for testing.
Mean and standard deviation is computed for all the images in the training data.
Training and test data is then normalized by subtracting this mean from each image and dividing each image by the computed standard deviation.
The network was of size 1024$-$512$-$1024 in size.
Adam~\cite{adam2014} was used as the optimizer to verify the validity of the proposed technique for different learning methods.
Parameters used for Adam were, $\alpha=0.001$, $\beta_1=0.9$, $\beta_2=0.999$ and $\epsilon=10^{-8}$.
The termination criteria was number of epochs that was set at $100$. Batch size was selected as $32$.

\begin{figure}[b!]
\centering
\begin{minipage}{.48\columnwidth}
  \centering
  \includegraphics[width=1.0\linewidth]{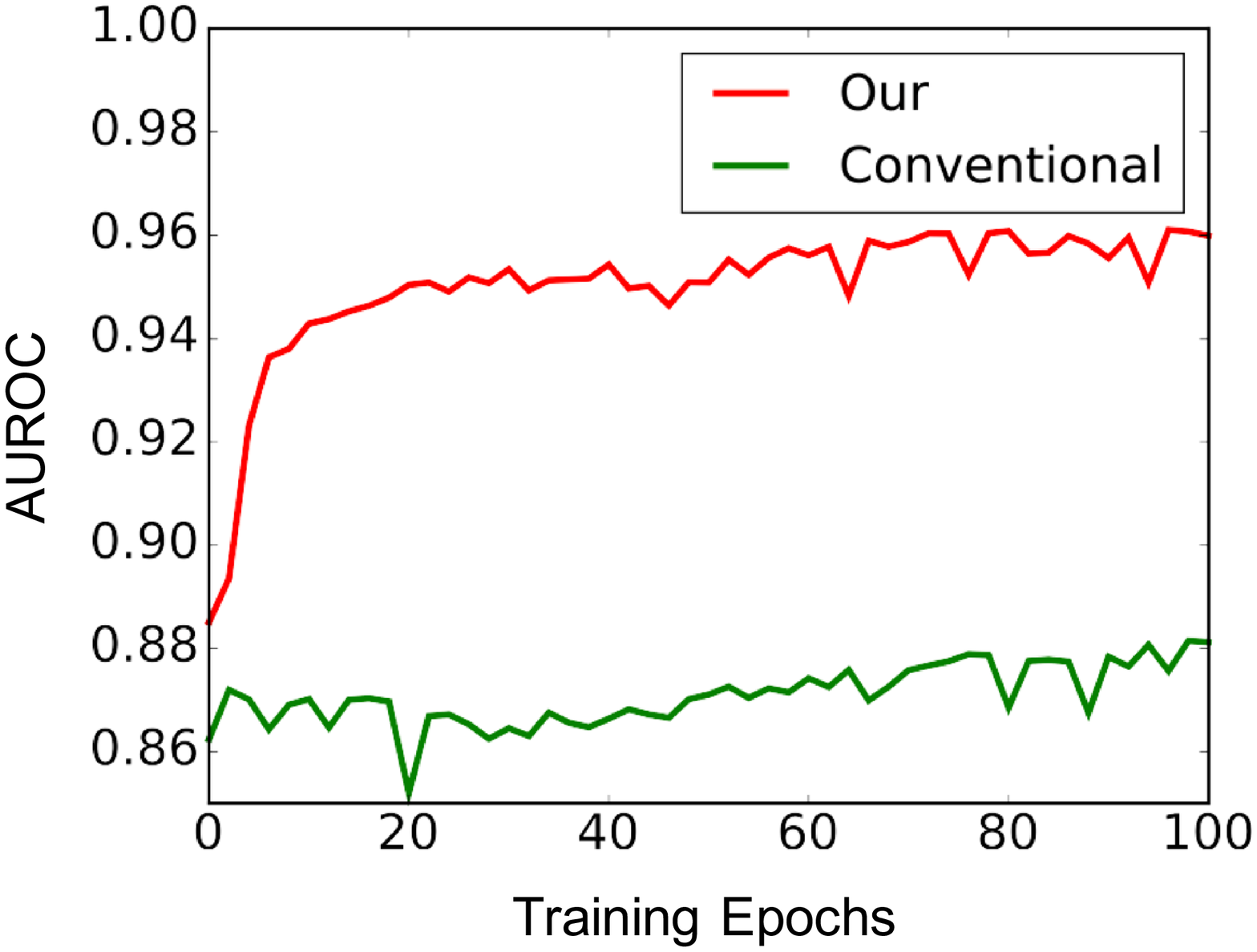}
  \captionof{figure}{AUROC convergence of the proposed method vs. conventional training of autoencoders.}
  \label{fig:auroc}
\end{minipage}%
\hspace{.03\columnwidth}
\begin{minipage}{.48\columnwidth}
  \centering
  \includegraphics[width=1.0\linewidth]{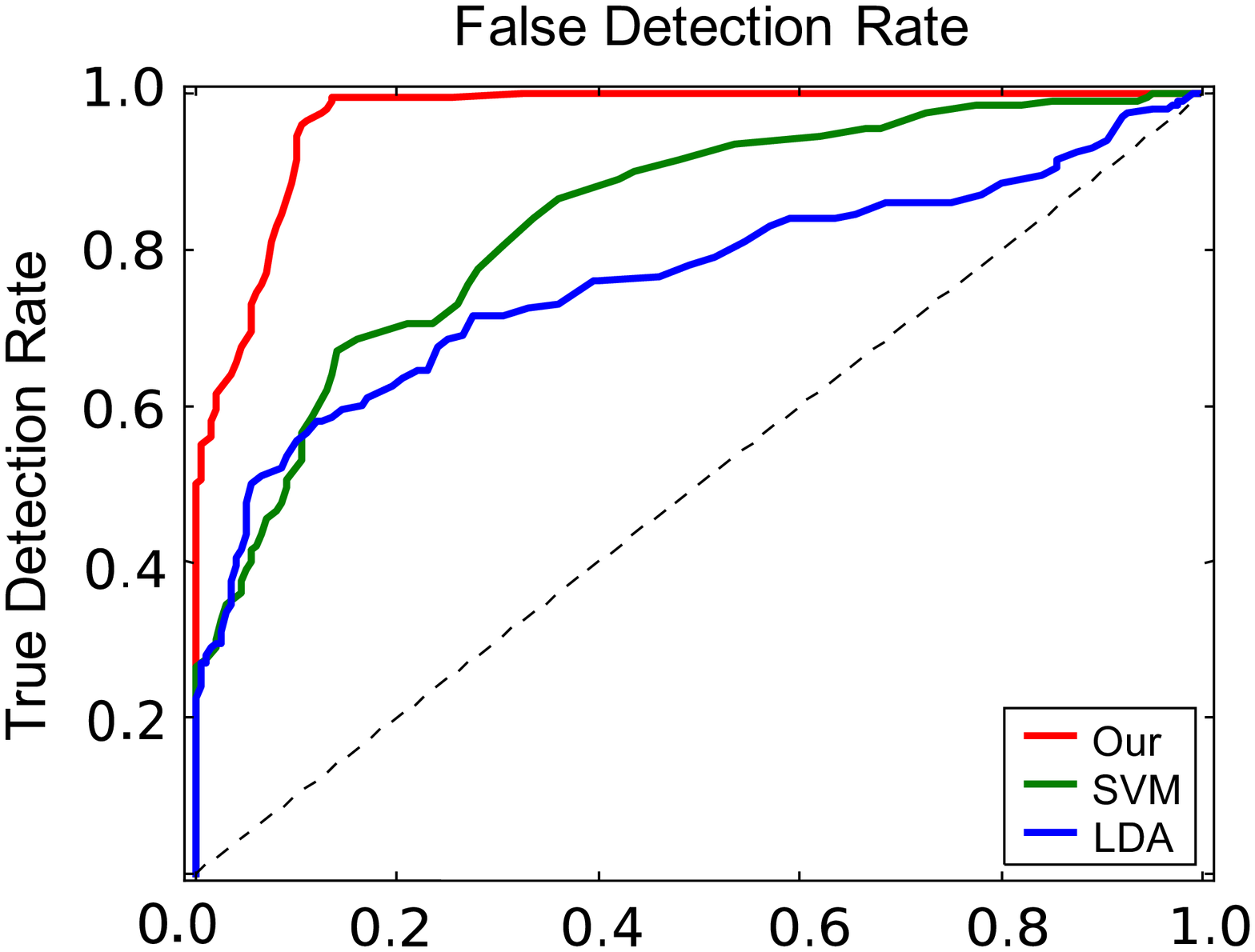}
  \captionof{figure}{ROC for simple two class classifiers vs. the proposed anomaly detection system.}
  \label{fig:roc}
\end{minipage}
\end{figure}

Area under the receivers operating characteristics curve (AUROC) was used for quantitative measure.
Figure~\ref{fig:auroc} shows that with conventional training of autoencoder using only the road data can still reconstruct the CIFAR images fairly well.
The value of AUROC for conventional autoencoders stays around $0.875$.
As the system learns to reconstruct the road better it becomes equally good at reconstructing the CIFAR data.
For the proposed technique the AUROC tends to reach $0.96$ as the number of epochs increases (for the proposed method one epoch means one iteration of positive and negative learning).
For this experiment the amount of negative learning to perform was computed adaptively by observing the maximum gain in AUROC by increasing or decreasing the size of data and iterations for negative learning.
Details on adaptive control of negative learning are out of the scope of this paper.
Figure~\ref{fig:reconstruction_roadcifar} shows the reconstruction quality for the road and CIFAR images by the conventional and the proposed approach.
It is clear that after $100$ epochs the conventional method can reconstruct the anomaly images much better then the proposed approach.
This results in lowering the AUROC for the conventional approach.

In Figure~\ref{fig:roc} we have compared the ROC of our system with classical two class classifiers. 
In this case a mask was used to locate the road in the video and anything outside the mask was treated as anomaly.
We captured a video on Japan highways in similar conditions to \cite{youtube1}.
However, while \cite{youtube1} use a relative wide angle 12-35mm lens, we used 70-150mm lens.
$50,000$ road patches of size $16\times16$ were extracted, while only $1,000$ patches were extracted for anomaly data.
The size of anomaly data was $50$ times less than that of the road data.
$70\%$ data was used as training and the remaining for testing.
The data was normalized in a manner similar to the previous experiment.
The AUROC for SVM and LDA classifier is $0.8326$ and $0.7526$ respectively.
The AUROC for the anomaly detection technique proposed in this paper reaches $0.9636$ in $100$ epochs.
The network was of size 256$-$200$-$256. The batch size was kept at $32$.
Vanilla stochastic gradient descent algorithm was used with a learning rate of $0.001$.
A significant improvement in AUROC clearly shows the benefit of the proposed approach.


\subsection{Limitations}
This technique works for the cases where non-anomalous data compared to anomaly data is confined in the input space.
In the above experiment treating CIFAR as normal and road as anomaly will not produce expected results.
This assumption is generally true for anomaly detection applications where normal operation is more or less predictable and uniform.

\section{Conclusions}
\label{sec:conclusions}

We proposed a novel method to train generative models namely autoencoders for anomaly detection.
Anomaly is determined by considering the similarity measure of the input and the reconstructed signal.
Conventional training methods allow the reconstruction of the signal space far beyond the training data.
The proposed method ensures that the autoencoder only learns to reconstruct the signals that are similar to the training distribution.
This makes it easier to separate a normal signal from an anomalous signal.
The core idea of this research is the introduction of a negative learning phase, in which the system unlearns the reconstruction of anomalous signals.
The balance of positive learning with negative learning phases help to move the frequency distribution curves for dissimilarity of regular and anomalous data away from each other.

As for a future direction, 
we are currently working on adding the notion of time to this approach.
In this case the system will only be able to predict the road feature that should appear next.
By matching prediction with the actual observation we can reveal anomalies.


\end{document}